%% file: main.tex
\documentclass{article}
\usepackage[nonatbib, preprint]{styles/neurips_2025}

\usepackage[utf8]{inputenc} 
\usepackage[T1]{fontenc}    
\usepackage{url}            
\usepackage{booktabs}       
\usepackage{amsfonts}       
\usepackage{nicefrac}       
\usepackage{microtype}      
\usepackage{xspace}
\usepackage{amsmath}
\usepackage{amssymb}
\usepackage{amsthm}
\usepackage{bbm}
\usepackage{bm}
\usepackage{caption}
\usepackage{dsfont}
\usepackage{enumitem}
\usepackage{float}
\usepackage{graphicx}
\usepackage{listings}
\usepackage[numbers]{natbib}
\usepackage{setspace}
\usepackage{subcaption}
\PassOptionsToPackage{dvipsnames}{xcolor}
\usepackage{tcolorbox}

\usepackage[bookmarks=false]{hyperref}
\hypersetup{
  colorlinks,
  linktocpage=true,
  urlcolor=Green,
  linkcolor=Green,
  citecolor=Green
}
\usepackage[capitalize,noabbrev]{cleveref}

\usepackage{balance}
\usepackage{dirtytalk}
\usepackage{tabularx}
\usepackage{comment}
\usepackage{tikz}
\usepackage{verbatim}
\usepackage{ragged2e}
\usepackage{multirow}
\theoremstyle{definition}
\newtheorem{definition}{Definition}

\newcommand{\gemma}{Gemma-2-2B\xspace}

\newcommand\extrafootertext[1]{%
    \bgroup
    \renewcommand\thefootnote{\fnsymbol{footnote}}%
    \renewcommand\thempfootnote{\fnsymbol{mpfootnote}}%
    \footnotetext[0]{#1}%
    \egroup
}

\newcommand*\samethanks[1][\value{footnote}]{\footnotemark[#1]}

\title{Detecting and Characterizing Planning \\ in Language Models}

\author{
Jatin Nainani \\
University of Massachusetts Amherst\\
\texttt{jnainani@umass.edu} \\
\And
Sankaran Vaidyanathan \\
University of Massachusetts Amherst\\
\texttt{sankaranv@cs.umass.edu} \\
\And
Connor Watts \\
Queen Mary University of London\\
\texttt{c.watts@qmul.ac.uk} \\
\And
Andre N. Assis\thanks{Equal Contribution} \\
Independent\\
\texttt{anogueira.assis@gmail.com} \\
\And
Alice Rigg\samethanks \\
Independent\\
\texttt{rigg.alice0@gmail.com} \\
}

\begin{document}
\maketitle

\urldef{\githublink}\url{https://github.com/ambitious-mechinterp/plan_trace}
\extrafootertext{Code available at \githublink}
\input{sections/abstract}
\input{sections/intro}
\input{sections/planning}

\input{sections/experiments}
\input{sections/base_vs_instruct}
\input{sections/discussion}


\bibliographystyle{unsrtnat}
\bibliography{sections/references} 


\clearpage
\onecolumn
\appendix

\input{appendices/background}

\input{appendices/case_study_additional}
\input{appendices/cantsaycases}
\input{appendices/ethics}
\input{appendices/artifacts}
\input{appendices/compute}
\input{appendices/aistatement}


\end{document}

%% file: sections/abstract.tex
\begin{abstract}

Modern large language models (LLMs) have demonstrated impressive performance across a wide range of multi-step reasoning tasks. Recent work suggests that LLMs may perform \textit{planning} — selecting a future target token in advance and generating intermediate tokens that lead towards it — rather than merely \textit{improvising} one token at a time. However, existing studies assume fixed planning horizons and often focus on single prompts or narrow domains. To distinguish planning from improvisation across models and tasks, we present formal and causally grounded criteria for detecting planning and operationalize them as a semi-automated annotation pipeline. We apply this pipeline to both base and instruction-tuned \gemma models on the MBPP code generation benchmark and a poem generation task where Claude 3.5 Haiku was previously shown to plan. Our findings show that planning is not universal: unlike Haiku, \gemma solves the same poem generation task through improvisation, and on MBPP it switches between planning and improvisation across similar tasks and even successive token predictions. We further show that instruction tuning refines existing planning behaviors in the base model rather than creating them from scratch. Together, these studies provide a reproducible and scalable foundation for mechanistic studies of planning in LLMs.

\end{abstract}

%% file: sections/intro.tex
\section{Introduction}
\label{sec:intro}

Large language models (LLMs) have achieved impressive results on complex reasoning tasks, from creative writing to code generation \cite{brown2020fewshot, odena2021programsynthesis}. These tasks often require multi-step reasoning, yet LLMs are trained as next-token predictors that would presumably generate outputs by \textit{improvising} each token step-by-step without foresight. An alternative hypothesis is that LLMs solve these tasks by \textit{planning}: using intentional processes, whether internal or external, that guide generation in a structured, goal-directed way to improve coherence and reasoning. This distinction is critical because planning may be necessary for reliable chain-of-thought reasoning and long-horizon problem solving, while hidden planning mechanisms could enable models to pursue unintended goals or conceal their reasoning. Therefore, it is essential to determine not just whether or not planning occurs, but also the \textit{mechanisms} through which planning arises. 

Recent work provides evidence that LLMs engage in internal planning. For example, researchers have probed for representations of future tokens at fixed distances ahead in language models \citep{Pal_2023, pochinkov2024extractingparagraphsllmtoken, wu2024do} and games like chess \citep{lc0, jenner2024evidencelearnedlookaheadchessplaying} and Sokoban \citep{bush2025interpretingemergentplanningmodelfree}. \citet{lindsey2025biology} showed that Claude 3.5 Haiku was shown to generate a poem by storing candidate rhyme words before writing the next line, and ablating the representations for these rhyme words alone can potentially change the entire next line. However, these studies have key limitations: they assume fixed planning horizons, need task-specific probes, or work only in narrow domains. Without unified and robust tools for distinguishing between planning and improvisation, it is difficult to systematically compare behaviors across architectures and prompts or understand how planning in LLMs fundamentally works.

In this work, we address these gaps by formalizing a general, falsifiable definition of planning that applies to arbitrary LLMs and tasks (\S \ref{sec:def_planning}). Our approach makes two technical advances. 
First, we translate the intuitive notion of planning into concrete, causally grounded criteria at the activation level and provide an operational pipeline for annotating a forward pass as planning or improvisation.
Second, we apply this pipeline to the planning in poems example from \citet{lindsey2025biology} as well as a subset of the MBPP \cite{odena2021programsynthesis} code generation benchmark. Our findings include the following: 

\begin{itemize}[leftmargin=2em]
    \item In contrast with \citet{lindsey2025biology}, who found clear evidence for backward planning in Claude on a rhyming task, we find that \gemma solves the same task successfully by improvisation, without explicit intermediate planning signals (\S \ref{sec:planning-in-poems}).
    \item When we generalized to the larger MBPP benchmark, we found that \gemma switches between planning and improvisation within tasks and even within successive token predictions. We also demonstrate the examples where our criteria do not have a sure answer (\S\ref{sec:cant_say}).
    \item In all cases where \gemma Instruct can solve an MBPP task, \gemma Base consistently solves the improvisation cases correctly but shows lower performance on cases where planning is involved. The base model still engages in planning, but executes it less effectively. (\S\ref{section:instruct-base}). We find evidence that instruction tuning refines planning behavior rather than creating it.

\end{itemize}

With these studies, alongside an explicit and reproducible pipeline for verifying planning and improvisation, we aim to scale up and advance mechanistic studies of LLM reasoning.

\begin{figure}[t]
    \centering \includegraphics[width=\linewidth]{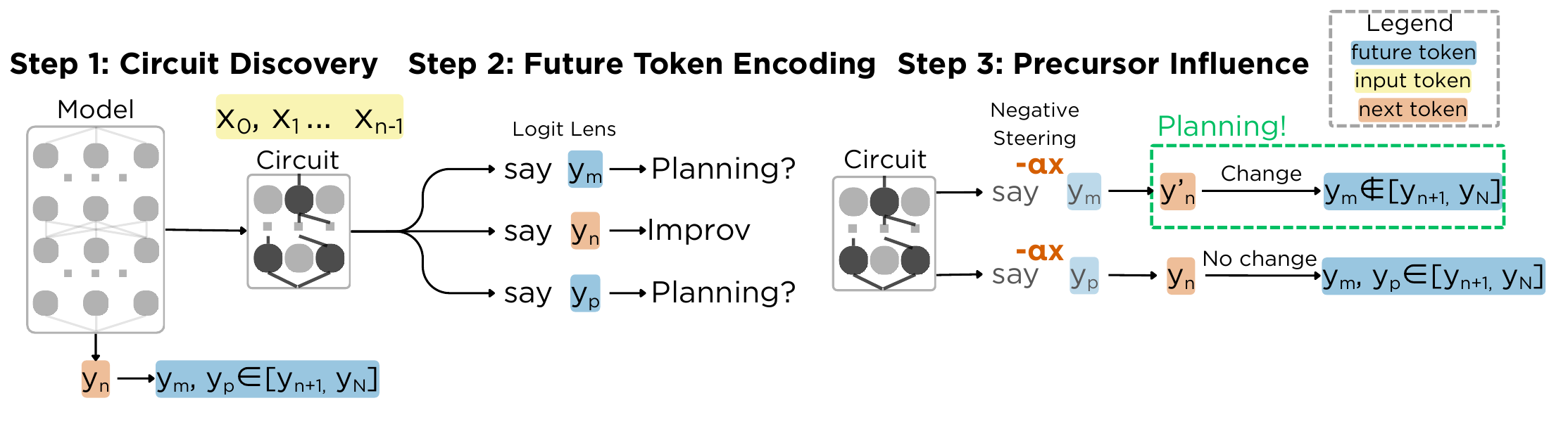}
\caption{Planning detection at a glance.
While predicting the next token $y_n$, we ask if the model is already planning for a later token $y_m$.
\emph{Step 1} isolates a smaller SAE-feature circuit that causes $y_n$.
\emph{Step 2} (Future-Token Encoding, FTE) uses a Logit-Lens readout to see which features ``write'' towards a future token (blue) versus the next token (red); future-writing features are planning candidates.
\emph{Step 3} (Precursor Influence, PI) negatively steers each candidate at its earliest use (orange $-\alpha x$). 
If this steering changes the future token $y_n$, perturbs intermediate tokens, and removes $y_m$ from the output, we call it \textsc{Planning}.}

    \label{fig:intro_fig}
\end{figure}

%% file: sections/planning.tex
\newcommand{\ym}{$y_m$}
\newcommand{\yn}{$y_n$}

\section{Defining and Detecting Planning}\label{sec:def_planning}

We aim to turn the intuitive idea of planning in LLMs into a general-purpose and empirically verifiable definition. This section introduces the formal criteria we use throughout the paper and an operational pipeline for verifying it on real datasets and models.

\subsection{Motivating Example}
\label{sec:motiv_eg}
Consider the task of completing a rhyming couplet as studied by \citet{lindsey2025biology}, where the model is given the following prompt:

\begin{center}
    \textit{A rhyming couplet:}\texttt{\textbackslash n} \textit{He saw a carrot and had to grab it,}\texttt{\textbackslash n}
\end{center}

In this task, the model is expected to return a line that rhymes with the end words ``grab it". To write the next line, the model could use either of the following strategies:

\begin{itemize}[leftmargin=2em]
\item \textbf{Improvisation.}  Generate each token one-by-one and choose a rhyming end word (e.g. ``rabbit'' and ``habit'') only at the final position.
\item \textbf{Planning.}  Decide on the rhyming word in advance (e.g. ``rabbit'') and generate each subsequent token to ensure the line ends with the chosen word.
\end{itemize}

\citet{lindsey2025biology} investigated a variant of the above prompt in Claude 3.5 Haiku \citep{anthropic2024claudehaiku} and demonstrated that it plans ahead for two possible rhyming words, ``rabbit" and ``habit". To do this, they showed that internal representations of the words ``rabbit" and ``habit" are active at the next position after the prompt. By default, the model returned a line ending with ``like a starving rabbit", but suppressing features associated with the word ``rabbit" led to the model returning ``a powerful habit'' instead. The fact that suppressing the ``rabbit" features led the model towards a different output indicates that those ``rabbit" features had a causal effect on the resulting output.

This intuition suggests that planning requires an internal representation of a future token that is active at an earlier position in the sequence and causally influences the generation of all subsequent tokens leading up to it. We formalize these two requirements below. 

\subsection{Formalizing Planning}

In this paper, we use sparse autoencoder (SAE) latents \citep{huben2024sparse} as interpretable representations of tokens and concepts in the model. We denote these representations through triples $(l,f,t)$, where $l$ is a layer index, $f$ is an SAE latent, and $t$ is a token position. Consider a prompt $(x_0,\dots,x_{n-1})$ and let $(y_n,\dots,y_N)$ be the output tokens generated by the model. We call $y_n$ the \textit{current} token and
$y_m$ ($m>n$) a \textit{future} token. 

\begin{definition}[Future‑Token Encoding]
\label{def:fte}
     Let $W_l[f]$ be the decoding direction for latent $f$ at layer $l$. For any candidate future token $y_m$ with $n < m \le N$, if $y_m$ appears in the top $K$ tokens when projecting $W_l[f]$ through the unembedding matrix, then $f$ is said to be a future-token encoding for $y_m$.
\end{definition}

\begin{definition}[Precursor Influence]
\label{def:pi}
    For some $\alpha > 0$ and token position $t$, if subtracting a scaled decoding direction $\alpha W_l[f]$ from the residual stream at $(l,t)$ during the forward pass and regenerating the sequence from $t+1$ causes
    \begin{enumerate}[label=(\roman*)]
    \itemsep0em
        \item a change in the next token $y_n$
        \item a change in at least one intermediate token $y_{n+1},\dots,y_{m-1}$
        \item removal of $y_m$ from the generated output
    \end{enumerate}
    then the latent $f$ has a precursor influence on the future token $y_m$. 
\end{definition}

\begin{definition}[Planning]
\label{def:planning}
    A model is \textbf{planning} at position $(l,t)$, during the prediction of $y_n$, for a future token $y_m$ ($m>n>t$) if there exists a latent $f$ at $(l,t)$ that is a future-token encoding (\textbf{FTE}) and has a precursor influence (\textbf{PI}) on the future token $y_m$.
\end{definition}

We are not proposing that these definitions are complete and exhaustive. They are built considering the following working assumptions, and knowing these help us describe the boundaries of where these definitions will work. 

\begin{enumerate}[leftmargin=2em]
\itemsep0.2em
    \item If a model is planning for \ym during the prediction of \yn, then the circuit for predicting \yn will have latents related to \ym.
    \item If \ym is in the top $K$ of logit lens \citep{nostalgebraist2020logitlens} for a latent, then the latent is increasing the logit probability for \ym or ``thinking'' about \ym and it is ``related to \ym''.
    \item If planning for a future token occurs, then intermediate tokens are affected. 
    \item Negative steering a latent suppresses the token/concept it is related to from the activation space. 
\end{enumerate}

\subsection{Feature Roles Induced by the Criteria}
\label{sec:feature_roles}

The \emph{Future‑Token Encoding} (\textbf{FTE}) and
\emph{Precursor Influence} (\textbf{PI}) criteria can be used to partition the set of
$(l,f,t)$ triples into the following behaviorally distinct classes:

\textbf{Planning:}
          A planning feature satisfies \emph{both} \textbf{FTE} and \textbf{PI} for some future token $y_m$ that is absent from the prompt.
          In other words, it stores a representation of $y_m$ and exerts an early causal influence that shapes the intermediate trajectory towards $y_m$. Removing the feature at the point where it first activated will prevent the token $y_m$ from being generated, and usually steers the generation down a semantically unrelated path.

\textbf{Improvisation:}
      A feature satisfies \textbf{FTE} for some $y_m$ but not \textbf{PI}; it only exerts a causal influence at the position just before $y_m$ is generated. In other words, steering or ablating that feature right before $y_m$ can change the next token, but doing the same at an earlier position will not change $y_m$ or any of the intermediate tokens leading up to it. 

\textbf{Neither:}
          A feature fails \textbf{FTE} for every future token. This does not mean that the feature is not important; it could be encoding computations that keep the language model ``on track'' without explicitly referencing a future goal. These can include local syntax, formatting, and short‑range semantics,
          discourse markers, duplicate‑token detectors, or many others.
          
\textbf{Can't Say:} This category represents scenarios that could be ambiguous, meaning the existence of a causal effect need not be interpreted as planning behavior.

\begin{itemize}[leftmargin=2em, nosep]
        \item \textit{Overlap with Prompt:} The goal token $y_m$ already
        appears earlier or ties another future token in the Logit Lens ranking. Even if \textbf{FTE} is satisfied, it is unclear whether this is just attending to a token in the prompt or planning for a future token.
        
        \item \textit{Out‑of‑Distribution Steering:} When steering at an earlier position results in degenerate or nonsensical outputs, it is unclear whether to consider this to be the same as suppressing the planning mechanism. Hence, even though the feature technically satisfies \textit{PI}, we do not label it as a planning feature.
    \end{itemize}
    Appendix \ref{appendix:cant-say} includes examples and potential strategies for identifying planning and improvisation in these scenarios. For this work, we exclude these cases from all quantitative metrics. 

\subsection{Identifying Planning at Scale}
\label{section:operational-pipeline}

For any model and prompt, we could apply Definition \ref{def:planning} to every $(l,f,t)$ triple and label them as \textsc{Plan} or \textsc{Improv}. However, this is usually infeasible in practice. For example, Gemma-2‑2B would require \( 26\text{ layers}\times16K\text{ latents}\times100\text{ tokens} \approx4.2\times10^{6} \) tests per prompt. Indeed, many prior works on planning in LLMs focus on a single prompt or a handful of prompts. 

We provide a pipeline to trim the search space by a factor of \(\sim\!10^{4}\) while preserving almost all genuine planning positions. An overview of our detection pipeline (circuit discovery to FTE to PI) is shown in Fig.~\ref{fig:intro_fig}.

\begin{enumerate}[label=\textbf{Step~\arabic*}, start=0, nosep] 
\item \textbf{Circuit Discovery:} 
      Because \textbf{PI} already requires a causal effect on the next token $y_n$, we first isolate the sparse feature circuit that \emph{explains} the prediction of $y_n$. Starting with the latents that have the highest indirect causal effect, we build the
      smallest set of $(l,f,t)$ triples $\mathcal{C}$ that can recover the original logit distribution $P_{\text{model}}(y_n)$ by at least \(60\%\) when all other $(l,f,t)$ triples are zero-ablated. Empirically we find that $\lvert\mathcal{C}\rvert\in[2\times10^{4},3\times10^{4}]$, which is represents a decrease by a factor of $150\times$.
    
\item \textbf{Future‑Token Encoding Filter: } 
      Apply \textbf{FTE} to every triple $(l,f,t)\in\mathcal{C}$ with $t<n$.
      Keep a triple only if its Logit‑Lens top‑$K$ contains some future
      token $y_m$, otherwise label it as \textsc{Neither}.
      Triples that share the same $(l,t)$ and point to the same $y_m$
      are merged into a \textit{cluster} $S$. In our experiments, each cluster contained on average $\sim\!50$ $(l,f,t)$ triples.
\item \textbf{Cluster‑Level Precursor Influence Check: }
      Steering the whole cluster at once is \(\sim\!50\times\)
      cheaper than steering its individual members. We subtract
      $\alpha\!\sum_{(l,f,t)\in S}\!W_l[f]$ when predicting $y_n$ for a range of $\alpha$ values. If \textbf{PI} is satisfied for the target $y_m$ without a degenerate output, the cluster is kept as a \textsc{Plan} candidate; otherwise, the whole cluster is considered a candidate for \textsc{Improv} or \textsc{Can't say} for nonsensical generations. Note that all $(l,f,t)$ triples inside $S$ satisfy \textbf{FTE} by construction, so
      \textsc{Neither} cannot occur here. 
\item \textbf{Earliest‑Moment Search:} For within surviving clusters, greedily walk backwards through the positions where $S$ is active, ablating one triple at a time until PI fails. The last triple whose removal still deletes $y_m$ is recorded as the first backward‑planning moment. Other $(l,f,t)$ triples in $S$ that satisfy both \textbf{FTE} and \textbf{PI} are also labeled \textsc{Plan}.

\item \textbf{Improvisation Check:} We rerun Step 2 for all $(l,f,t)$ that are not already labeled as \textsc{Plan} but with $y_m$ as the next token. Any $(l,f,t)$ triple that has a causal effect on $y_m$ without satisfying \textbf{PI} for any of the previous tokens is labeled as \textsc{Improv}. For all $y_m$ already present in the input prompt, we also assign \textsc{Can't Say}. The remaining are labeled as \textsc{Neither}.

\end{enumerate}

Step 0 focuses solely on $y_n$ because any feature that fails to influence the next token being predicted \emph{cannot} satisfy \textbf{PI}. Following from \citet{lindsey2025biology}, we use cluster‑level steering in Step 2 to amortize
compute since discarding even two clusters early saves $\sim100$ individual \textbf{PI} checks later. In the next section, we will empirically evaluate the above pipeline on real-world data.

%% file: sections/experiments.tex
\section{Empirical Evaluation}
\label{sec:empirical-findings}

We empirically evaluate our detection framework on the \textsc{Base} and \textsc{Instruct} versions of \gemma \citep{gemmateam2024gemma2improvingopen}. We used TopK SAEs trained on \texttt{MLP\_out} from the GemmaScope suite \citep{lieberum2024gemmascopeopensparse}. These SAEs are trained on the outputs of each MLP block before RMSNorm is applied. 

Our analysis consists of three main components. We first provide a motivating example of our criteria/pipeline on several rhyming‑couplet tasks (\S\ref{sec:planning-in-poems}) to give a direct comparison to prior work \citet{lindsey2025biology}. We then demonstrate our detection framework on several programming tasks (\S\ref{sec:planning-in-code}). Finally, we provide a comparative analysis of planning in \textsc{Base} vs. \textsc{Instruct} models (\S\ref{section:instruct-base}).

\subsection{Planning in Poems}
\label{sec:planning-in-poems}

We now revisit the example in \S\ref{sec:motiv_eg} to evaluate our criteria and pipeline. \citet{lindsey2025biology} showed that Claude 3.5 Haiku activates latent features for candidate rhyme words such as \textit{habit} and \textit{rabbit} at the end of the first line (\texttt{'\textbackslash n'}), six tokens before the model predicts the second rhyme - \textit{``rabbit''}.
When the same prompt, \textit{``A rhyming couplet:}\texttt{\textbackslash n} He saw a carrot and had to grab it, \texttt{\textbackslash n''}, is given to \gemma \textsc{Instruct}, it completes it with \textit{``A tasty treat, a crunchy habit.''} 

Running our FTE \(+\) PI pipeline (\S\ref{section:operational-pipeline}) over every intermediate prediction (from \(y_{1}\) = "A" through \(y_{6}\) = "tasty") reveals that \emph{no circuit satisfies both criteria}. In other words, Gemma shows \emph{no evidence of planning} during this poem generation.
A binary verdict alone does not illuminate the model’s internal strategy, so we manually inspected the full circuit for predicting ``habit''(\(\sim 26\text{k}\) latents, \(\geq 60\%\) logit recovery), similar to the setup of \citet{lindsey2025biology}. 
The circuit contains two distinct groups of latents: one that activates on phoneme-level tokens (e.g.~``/t/'', ``/et/''), and another that activates on compulsion tokens(e.g.~``had to grab'', ``must''). Latents writing to ``habit'' only become causally relevant at the \emph{final} token. \citet{lindsey2025biology} also demonstrated that negative steering on the ``habit'' latents caused a change in the intermediate tokens, which didn't happen for this circuit in Gemma. Thus, our working hypothesis is one of \emph{improvisation}: local phonetic and thematic cues combine late to select the rhyme, rather than a plan propagated forward from line one.

Differences in planning are expected given variations in architecture, scale, and training data. Our semi-automated pipeline surfaces those discrepancies, providing a systematic lens for future work on how modeling choices shape emergent planning behavior.
We now move to evaluating the pipeline on code generation tasks, as coding tasks are well-represented in the training data for Gemma 2 \citep{gemma2_modelcard_2025}. 

\subsection{Planning in Code}
\label{sec:planning-in-code}

We next execute and analyze the detection framework on several programming tasks. For this we consider the Mostly Basic Programming Problems (MBPP) dataset \citep{austin2021programsynthesislargelanguage}. 
We filtered the tasks to include only those that the \textsc{Instruct} model solves correctly, picking the first 60 for analysis. We then run the pipeline on this set of tasks. In the following, we provide a selection of case studies where the model exhibits planning by our criteria.

\begin{figure}[t]
    \centering \includegraphics[width=\linewidth]{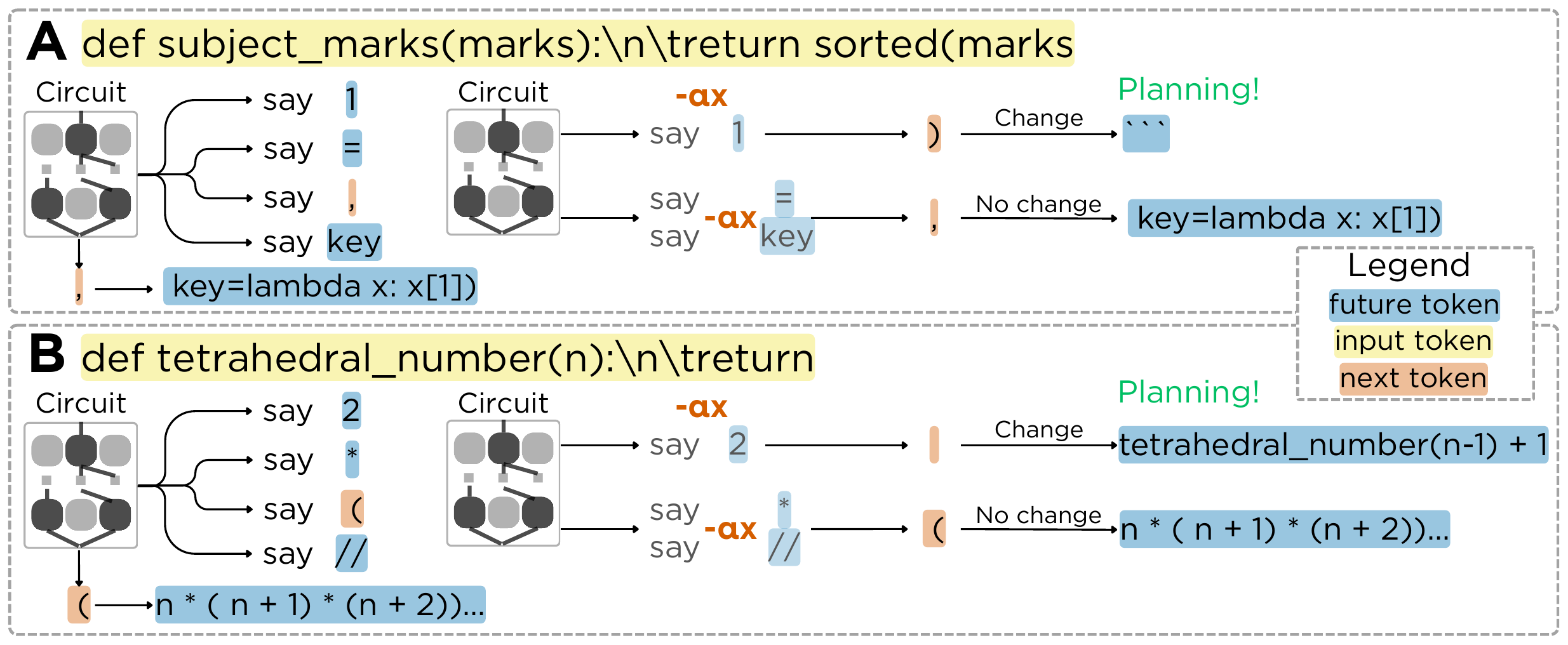}
\caption{Planning in MBPP tasks.
\textbf{(A) Sort tuples by the 2nd element:} While predicting the comma after ``\texttt{sorted(marks}'', a feature already promotes the future token \texttt{"1"} present in \texttt{key=lambda x:\,x[1]} (FTE). Suppressing it changes the generation to close the bracket instead (PI).
\textbf{(B) $n$-th tetrahedral number:} While predicting the first parenthesis of the closed form, a feature encodes the later \texttt{"2"} in $(n{+}2)$ (FTE). Suppressing removes the plan: the model drifts to a recursive sketch (PI).}
    \label{fig:case_studies_2}
\end{figure}

\subsubsection{Sorting a list of tuples.}\label{sec:sortinglist}

This task involves sorting a list of tuples \verb|subjectmarks| \emph{in‑place} by the second element of each tuple. The \textsc{Instruct} model correctly solves the task by using a lambda function to key into the index 1 of the tuple for sorting:
\verb|subjectmarks.sort(key=lambda x: x[1])|.

\emph{Planning evidence.}  
During prediction of the comma ($y_n=297$), the model is already planning for the future token \texttt{"1"} ($y_m=305$) as early as $(\ell{=}0,\;t{=}294)$ \texttt{"sorted"}. The earliest feature responsible writes in the direction of \texttt{"1"}, ranking it first among top‑10 logits (\textsc{FTE}). Suppressing (negatively steering) this feature flips the next token from the comma to a closing parenthesis (\emph{i}). The generation just adds a newline and ends the function (\emph{ii}), and \texttt{"1"} never appears in the continuation (\emph{iii}). More intuitively, the model outputs the comma because it is planning to emit \texttt{"1"} later (it must sort by the second element of each tuple). Suppressing \texttt{"1"} features at this position causes the model to just close the bracket. The steered generation fails the unit test. See Fig.~\ref{fig:case_studies_2}A for a schematic of this example (Appendix \S\ref{sec:app_sort}).

\subsubsection{Computing the $n$‑th tetrahedral number.} \label{sec:tetrahedralcase}
This task involves computing the $n$-th tetrahedral number $T(n)$. The \textsc{Instruct} model correctly solves the task by using the closed form \verb|(n*(n+1)*(n+2))//6| after handling small $n$, satisfying the tests. 

\emph{Planning evidence.}  
During prediction of the first opening bracket of the tetrahedral number formula ($y_n{=}180$), the model is already planning for the future token \texttt{"2"} as early as $(\ell{=}0,\;t{=}18)$ (\texttt{"find"}). The earliest feature responsible writes in the direction of \texttt{"2"}, ranking it among the top-10 logits (\textsc{FTE}). Suppressing (negative steering) this feature switches the model from the closed form to the recursive update \verb|tetrahedral_number(n-1)+1|: (\emph{i}) \texttt{"tet"} is predicted as the next token instead of the opening parenthesis, (\emph{ii}) the predictions after \texttt{"tet"} complete the recursive call, and (\emph{iii}) \texttt{"2"} never appears (\textsc{PI}). This generation fails the unit tests. Intuitively, the model places the parenthesis because it is planning to emit the \texttt{"2"} needed for the $\times(n+2)$ factor; removing that plan pushes it back to a simpler recursive sketch that does not pass the tests. See Fig.~\ref{fig:case_studies_2}\textbf{B} for the corresponding schematic (Appendix \S\ref{sec:app_tetra}).

\subsubsection{Forming the maximum number from digits.}\label{sec:formmax}
Given a list of digits, the task is to return the largest possible integer formed by concatenating all elements from the list. The \textsc{Instruct} model correctly solves the task by sorting the list in descending order and traverses it with a for‑loop whose index variable is \texttt{i}.

\emph{Planning evidence.}  
During prediction of \texttt{"digits"} (the first non-docstring token, $y_n{=}191$), the model is already planning for the future token \texttt{"sort"} as early as $(\ell{=}17,\;t{=}190)$, which is the first tab (\texttt{"\textbackslash t"}) after the docstring. The earliest feature responsible writes toward \texttt{"sort"}, placing it in the top-10 logits (\textsc{FTE}). Suppressing (or negatively steering) the \texttt{"sort"} feature at the \texttt{"digits"} token position (\emph{i}) flips the next token to \texttt{max}, (\emph{ii}) yields a program that instead begins \verb|max_num =| and fails all hidden tests, and (\emph{iii}) removes \texttt{"sort"} entirely from the continuation (\textsc{PI}). Intuitively, the model commits to \texttt{"digits"} because it is planning to immediately call \verb|sort|; without that plan, it never orders the digits and thus cannot construct the maximum number. See Fig.~\ref{fig:case_studies_3rd} for the schematic (Appendix \S\ref{sec:app_max_num}).

\subsubsection{Examples of ``can't say'' cases.}
\label{sec:cant_say}
\paragraph{Divisible tuples} The task is to return only those tuples whose elements are all divisible by a given number $k$. The baseline generation correctly completes the comprehension \texttt{"== 0 for element in tup)"} and appends matching tuples before returning, satisfying the tests. For the prediction of the next token \texttt{"=="}, we find features writing to the direction of \texttt{"for"}, which is a future token. With the steering token \texttt{"for"} (coeff $-80$), the output collapses around the divisor check into something like \texttt{", k"):} and loses the generator expression, yielding a syntactically invalid snippet. This satisfies both FTE and PI, but because \texttt{for} is also in the input, we label it as ``can't say'' (Appendix \S\ref{sec:op_prompt}).

\paragraph{Largest number from digits} The task is to rearrange a list of digits to form the maximum possible integer. The baseline sorts the list and builds the number by concatenating digits in reverse order (e.g., \texttt{max\_num += str(digits[n-i-1])}) and returns \texttt{int(max\_num)}, which passes the tests. For the prediction of the next token \texttt{"num"}, we find features writing to \texttt{"-"}, which is a future token and is not in the input. Under steering with the token \texttt{"-"} (coeff $-80$), the model veers into nonsense (e.g., \texttt{digits = len(digits)} followed by stray triple-quoted lines), so the steered output is degenerate (Appendix \S\ref{sec:ood_steer}).

\paragraph{Tetrahedral number} This is the same task as \S\ref{sec:tetrahedralcase}, but for a different forward pass. With \texttt{else} being the next token. 

With the steering token \texttt{"("} (coeff $-100$), the generation devolves into a stream of ``the/The'' without code or logic, so the steered output is degenerate. Note that the steered token \texttt{"("} is in the original prompt already, in the function signature and assertions. (Appendix \S\ref{sec:op_ood}).

\smallskip
Across all three case studies, the same pattern emerges: an SAE‑cluster that (i) linearly encodes a distant goal token and (ii) causally steers multiple intermediate tokens is necessary for the model’s success, thereby validating our planning labels. Overall, our pipeline identifies the \textsc{Instruct} model as either planning or improvising on 24 out of 60 tasks.

%% file: sections/base_vs_instruct.tex
\subsection{Comparing Base and Instruction-Tuned Models}
\label{section:instruct-base}

Although base models are trained to predict the next token, post-training methods such as instruction tuning and RL introduce multi-step or goal-oriented objectives, and we hypothesize that these post-training methods result in stronger planning behavior. In this section, we explore this hypothesis and compare the planning behaviors between the \textsc{Instruct} and \textsc{Base} models.

\begin{table}[htbp]
  \centering
  \caption{Pass rates by task subset for the \textsc{Instruct} and \textsc{Base} \gemma models. Values are percentages; \(n\) denotes the number of tasks evaluated in each subset.}
  \label{tab:instruct-vs-base}
  \begin{tabular}{lcc}
    \hline
    \textbf{Task subset} & \textbf{\textsc{Instruct} model} & \textbf{\textsc{Base} model} \\
    \hline
    Planning tasks (\(n=13\))       & 100\% & 54\%  \\
    Improvisation tasks (\(n=11\))  & 100\% & 100\% \\
    \hline
  \end{tabular}
\end{table}

For this comparison, we focus on the 24 MBPP tasks identified in Section \ref{sec:planning-in-code} and described in Table \ref{tab:instruct-vs-base}, where our detection pipeline classified whether the \textsc{Instruct} model was planning or improvising. We evaluate the \textsc{Base} model on our chosen subset of MBPP that the \textsc{Instruct} model solves, and compare the performance on the \textsc{Improv} cases vs. the \textsc{Plan} cases.

We find that \textsc{Base} solves all tasks where \textsc{Instruct} was improvising, but only 54\% (7 out of 13) of tasks where \textsc{Instruct} was planning.  The \textsc{Base} model's ability to solve these planning tasks suggests two possibilities: either \textsc{Base} already possesses some planning capabilities without instruction tuning, or it can solve these tasks without planning. 

To examine these hypotheses, we apply our planning detection pipeline to the \textsc{Base} model on the same planning tasks. We find that in many cases \textsc{Base} is still capable of planning, but there are distinct failure modes that result in incorrect answers. We identify two primary failure patterns in the following subsections.

\begin{figure}[t]
    \centering
    \includegraphics[width=\linewidth]{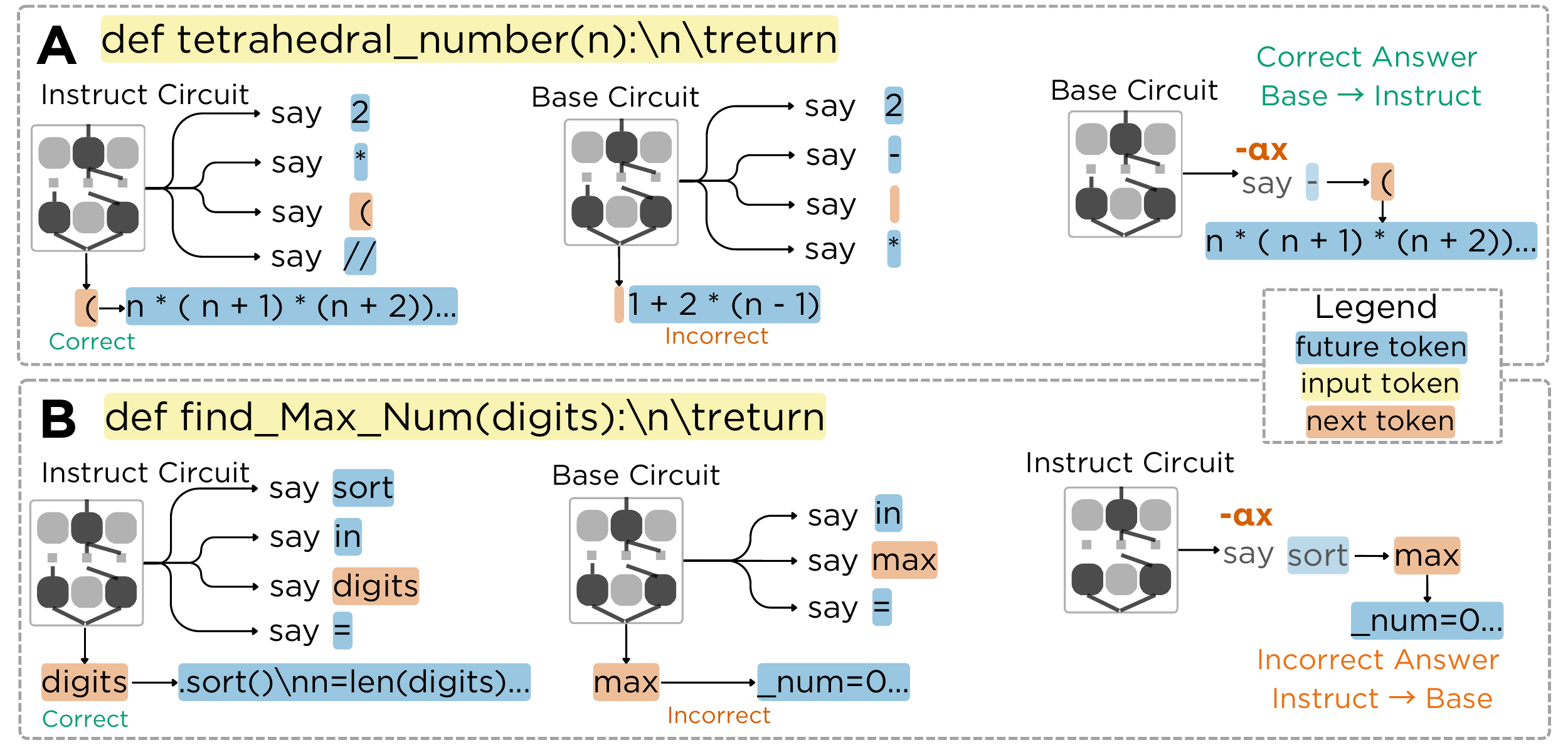}
\caption{Instruction tuning refines plan selection.
\textbf{(A) Competing plans (Tetrahedral number)}: Both models plan toward \texttt{"2"}, but \textsc{Base} also plans toward an alternative \texttt{"-"} path that yields an incorrect closed form. Suppressing \texttt{"-"} for \textsc{Base} removes the competing plan and recovers the correct solution, matching \textsc{Instruct}.
\textbf{(B) Incorrect Target (Largest number)}: \textsc{Instruct} plans to \texttt{sort} the \texttt{digits}, but \textsc{Base} plans toward \texttt{max} and never sorts, leading to failure. Suppressing \textsc{Instruct}'s \texttt{sort} plan reproduces the failed \textsc{Base} trajectory.
Diagrams highlight planning features; $-\alpha x$ indicates steering.}
    \label{fig:base_int_task24}
\end{figure}

\subsubsection{Competing Plans}

In the Nth Tetrahedral Number task covered in \S\ref{sec:tetrahedralcase} (Fig.~\ref{fig:base_int_task24}A, Appendix \S\ref{sec:app_tetra}), \textsc{Base} exhibits planning by targeting the correct token \texttt{"2"}, but also plans for an incorrect alternative \texttt{"-"}. When the model follows the incorrect plan, it generates the wrong formula (\texttt{1+2(n-1)}). In contrast, \textsc{Instruct} focuses solely on the correct plan and produces the right solution (\texttt{n*(n+1)*(n+2)/6}). 

Given these results, one potential explanation is that \textit{the \textsc{Base} model is still planning, but the plan is not yet specific enough}. If this were true, suppressing the wrong token that the model is also planning for should bring the behavior of \textsc{Base} closer to that of \textsc{Instruct}. Indeed, we find that suppressing the \texttt{"-"} token features causes \textsc{Base} to return the correct formula.

\subsubsection{Incorrect Target}

In the task for forming largest number from list covered in \S\ref{sec:formmax} (Fig.~\ref{fig:base_int_task24}B, Appendix \S\ref{sec:app_max_num}), \textsc{Base} is planning for the \texttt{"max"} token unlike \textsc{Instruct} which is planning for the \texttt{"digits"} token. However, unlike in the previous example, \textsc{Base} does not have \texttt{"digits"} as a potential plan, and therefore fails to return a correct answer. Suppressing the \texttt{"digits"} feature in \textsc{Instruct} leads it to return the same incorrect answer that \textsc{Base} does. 

Overall, these results suggest that base models still exhibit planning behavior, and instruction tuning is likely not the source of planning per se. However, instruction tuning can improve performance on planning tasks by helping the model select the right tokens to target.

%% file: sections/discussion.tex
\section{Conclusion}
\label{sec:discussion}

In this work, we introduced a general, falsifiable definition of planning in language models that generalizes and extends insights from prior case studies. 
We operationalized this definition through two criteria, Future-Token Encoding (\textbf{FTE}) and Precursor Influence (\textbf{PI}), and implemented a semi-automated pipeline to detect them. Applying this framework to the base and instruction-tuned versions of Gemma-2-2B on various MBPP code generation tasks, we demonstrated that:

\begin{itemize}[leftmargin=2em]
\itemsep0.01em
    \item \textbf{Planning is not universal.} The model solves some tasks by improvising and others by planning, and we found no clear rule governing which strategy is used. Planning does not appear to be task-specific either; Gemma-2-2B improvises on a poem generation task where Claude 3.5 Haiku was shown to plan \citep{lindsey2025biology}, though both models still generated valid poems.
    \item \textbf{Planning can be done poorly.}
    We found cases where the model deliberately planned toward incorrect answers or selected incorrectly among multiple competing plans.
    \item \textbf{Instruction tuning refines planning behavior but does not create it.} Both base and instruction tuned models are capable of planning, but it is possible that instruction tuning helps with choosing between competing plans or filtering out incorrect plans.
\end{itemize}

\subsection{Limitations}

\paragraph{SAEs} Our analysis was conducted using SAEs trained on the Gemma-2-2B base model but applied to the instruction-tuned version of Gemma-2-2B. While this approach is supported by prior work \citep{kissane2024saes}, it may introduce some mismatch in representation. Furthermore, we focused only on MLP-attached SAEs, inspired by their interpretability in prior work \citep{lindsey2025biology}. That said, our detection pipeline and criteria are general and can be extended to other types of SAEs and representation spaces.

\paragraph{Polysemantic latents} We found cases where some latents satisfy both FTE and PI, but upon manual investigation, we see that only one of the top 10 tokens is a future token and the others seem unrelated to the task. This is probably because the latent is polysemantic. We could potentially mitigate this by requiring stricter criteria such as a minimum threshold on autointerp scores.

\paragraph{Scaling to larger models and broader datasets}  Our study focuses on the base and instruct versions of Gemma-2-2B, mainly due to the availability of SAEs for all layers \cite{lieberum2024gemmascopeopensparse}. We plan to apply our detection framework to larger models in the Gemma family to understand how planning capabilities emerge as a function of scale. Additionally, our experiments have thus far focused on the MBPP dataset and the poem generation prompt from \citet{lindsey2025biology}. Extending this analysis to more challenging and diverse benchmarks could reveal deeper insights into planning behavior.

\subsection{Future Work}

\paragraph{Understanding edge and ``can't say'' cases}
A significant portion of our effort was spent on ambiguous or edge cases, where labels could not be clearly assigned. More details on such cases can be found in Appendix \ref{appendix:cant-say}; investigating these further could refine our definition and improve detection.

\paragraph{Automating offline detection} We found cases where steered generations satisfied our criteria for Precursor Influence, but the generated text itself was degenerate and nonsensical. It is not clear to us if this is an instance of planning behavior switching off or if the intervention pushed the model out of distribution. Setting thresholds for repeating tokens and perplexity could potentially help resolve this.

\paragraph{Online detection} This paper focuses on offline detection, i.e., detecting planning after the sequence is generated.  with the knowledge of future tokens. However, we believe it is possible to extend our approach to detect planning at test time, where we have no knowledge of the future tokens. For example, at each token prediction, we can find latents that write to tokens that are not present in the input; these become candidates for planning.

%% file: appendices/background.tex
\section{Background}
\label{appendix:background}

We review the three ingredients our method builds on: sparse autoencoders, causal‑influence localization, and prior work on planning in neural networks.

\subsection{Sparse Autoencoders}
Sparse autoencoders (SAEs) have gained popularity as an unsupervised interpretability method for analyzing activations of large language models. An SAE generally consists of an encoder-decoder structure: the encoder transforms the original activations into a higher-dimensional but sparse latent representation, while the decoder reconstructs the original activations from this sparse representation. 

We use SAEs from the GemmaScope suite (\citet{lieberum2024gemmascopeopensparse}). From this suite, we used TopK SAEs trained on MLP\_out. These SAEs are trained on the outputs of each MLP block, before RMSNorm is applied.

\subsection{Causal influence: activation \& attribution patching}
\label{sec:causal}
We follow \citet{marks2024sparse} for notations and approximations for circuit discovery with SAEs. 

\paragraph{Indirect effects}
Following \citet{vig2020investigating, finlayson2021causal}, let $m$ be any scalar metric of the forward pass (e.g.\ $-\!\log P_\theta(y_n)$)
and let $a$ be an internal activation. For a \emph{clean / patch} input pair $(x_\mathrm{clean},x_\mathrm{patch})$
we measure the \emph{indirect effect} (IE) \citep{pearl2022direct} of $a$ on~$m$ as
\begin{equation}
\label{eq:ie}
\mathrm{IE}\!\bigl(m;a;x_\mathrm{clean},x_\mathrm{patch}\bigr)
 \;=\;
 m\!\bigl(x_\mathrm{clean}\,\bigl|\operatorname{do}(a{=}a_\mathrm{patch})\bigr)
 - m\!\bigl(x_\mathrm{clean}\bigr),
\end{equation}
where the do‑operator fixes $a$ to its value
$a_\mathrm{patch}$ taken from the patched run.

However, computing \eqref{eq:ie} for every candidate $a$ is expensive,
so we adopt two gradient‑based approximations. \emph{Attribution patching}
\citep{nanda2023attribution, syed2023attributionpatchingoutperformsautomated, kramar2024atp} linearizes IE with a first‑order Taylor expansion, needing only \emph{two} forward passes and \emph{one} back‑propagation:
\begin{equation}
\label{eq:ap}
\widehat{\mathrm{IE}}_{\!\text{AP}}
 = \bigl.\nabla_a m\bigr|_{a=a_\mathrm{clean}}
   \,\bigl(a_\mathrm{patch}-a_\mathrm{clean}\bigr).
\end{equation}

\textbf{Integrated gradients (IG)} \citep{sundararajan2017axiomatic} trades extra compute for a tighter fit. Using $N{=}10$ evenly spaced interpolation points $\alpha\!\in\![0,1]$
we form
\begin{equation}
\label{eq:ig}
\widehat{\mathrm{IE}}_{\!\text{IG}}
 = \frac1N
   \sum_{k=1}^{N}
   \Bigl.\nabla_a m\Bigr|_{a=a_\mathrm{clean}
   +\frac{k}{N}\bigl(a_\mathrm{patch}-a_\mathrm{clean}\bigr)}
   \,\bigl(a_\mathrm{patch}-a_\mathrm{clean}\bigr) ,
\end{equation}
which markedly improves accuracy.

\paragraph{Single‑prompt variant.}
When only one prompt is available we replace
$(x_\mathrm{clean},x_\mathrm{patch})$ with
$(x, x)$ and set $a_\mathrm{patch}=0$, i.e.\ we measure the drop
in $m$ under \emph{zero‑ablation} of $a$;
the same formulas \eqref{eq:ap}–\eqref{eq:ig} apply after
substituting $a_\mathrm{patch}\!\leftarrow\!0$.

\subsection{Planning in neural networks}
\label{subsec:planninglit}

\paragraph{Predicting future tokens (fixed $k$).}
Early work asked whether a single intermediate representation
linearly encodes the \emph{final} logits $k$ steps ahead.
\citet{Pal_2023} trained an affine probe that can predict the
top‑$k$ logits four tokens in the future in GPT‑2, but only for some
layers.
\citet{pochinkov2024extractingparagraphsllmtoken}\, extend this to
paragraph‑level topics, showing that the newline token between
paragraphs already carries topical information.
\citet{wu2024do} repeat the experiment across model scales and
find that small models exhibit little signal, whereas larger models
show modest top‑token predictability.
In all cases the horizon $k$ is fixed by the probe designer.

\paragraph{Learned look‑ahead in games and RL.}
Outside language, neural agents sometimes plan several moves ahead.
\citet{jenner2024evidencelearnedlookaheadchessplaying} detect
representations of optimal next moves up to three ply ahead in Leela
ChessZero \citep{lc0} by training chess‑specific linear heads.
\citet{bush2025interpretingemergentplanningmodelfree} identify
state vectors in a Sokoban‑playing agent that encode the sequence of
box moves needed to solve the puzzle, again with a fixed look‑ahead.
These studies reinforce the possibility of learned planning but remain
task‑specific and horizon‑bound.

\paragraph{Variable‑horizon planning.}
The poem‑rhyme case study of \citet{lindsey2025biology} shows that
a large language model (Claude 3.5 Haiku) stores candidate rhyme words
an arbitrary number of tokens in advance and that ablating this latent
collapses the rhyme. This example motivates the formal criteria we adopt in
Section~\ref{sec:def_planning}.

\paragraph{Gap addressed by our work.}  
All prior studies either assume a fixed horizon or require
task‑specific probes.  Our criteria work for any distance
$m-n\ge 1$ and rely only on model‑intrinsic SAEs plus causal
steering.

%% file: appendices/case_study_additional.tex
\section{Additional details for ``planning'' cases}\label{sec:add_cases}

This appendix presents three representative ``planning'' cases. For each case, we show:
\begin{itemize}
  \item the \textbf{Prompt Prefix} (truncated to the noted token),
  \item the \textbf{Baseline Generation} continuation,
  \item the \textbf{Steering Token} and its \textbf{Coefficient}, and
  \item the resulting \textbf{Steered Continuation}.
\end{itemize}
All snippets below are exact text captures.

\begin{figure}[t]
    \centering \includegraphics[width=\linewidth]{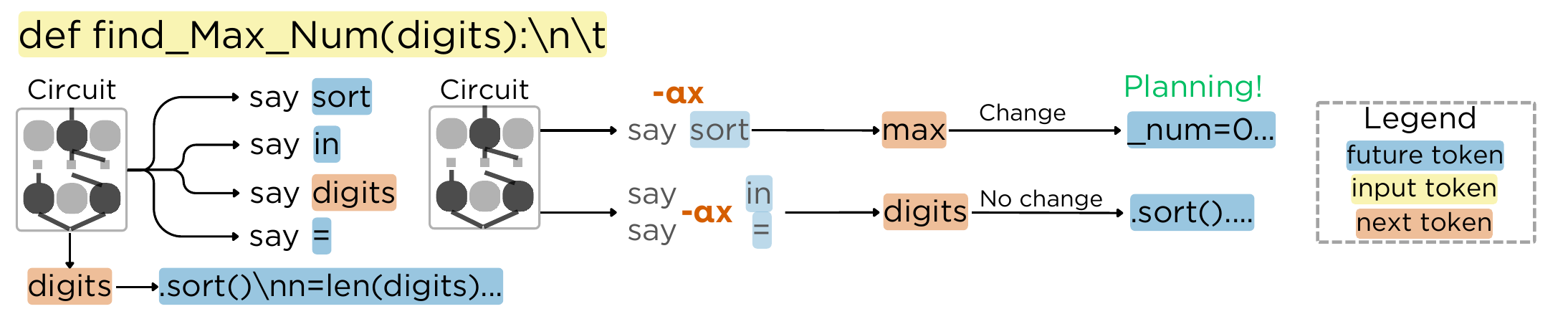}
\caption{Forming the max number from digits: While predicting the the first non-docstring ``\texttt{digits}'', a feature already promotes the future token \texttt{"sort"} present in \texttt{digits.sort()\textbackslash n} (FTE). Suppressing it changes the generation to not sort instead and start with \texttt{max\_num = 0} (PI).}
    \label{fig:case_studies_3rd}
\end{figure}

\subsection{Sorting list of tuples.}\label{sec:app_sort}

\textbf{Prompt Prefix (up to token 297):}

\begin{lstlisting}
<bos>You are an expert Python programmer, and here is your task: Write a function to sort a list of tuples using the second value of each tuple. Your code should pass these tests:

assert subject_marks([('English', 88), ('Science', 90), ('Maths', 97), ('Social sciences', 82)])==[('Social sciences', 82), ('English', 88), ('Science', 90), ('Maths', 97)]
assert subject_marks([('Telugu',49),('Hindhi',54),('Social',33)])==([('Social',33),
('Telugu',49),('Hindhi',54)])
assert subject_marks([('Physics',96),('Chemistry',97),('Biology',45)])==([('Biology',45),
('Physics',96),('Chemistry',97)])
Write your code below starting with "```python" and ending with "```".
```python
def subject_marks(marks):
    """
    Sorts a list of tuples by the second value of each tuple.

    Args:
        marks: A list of tuples, where each tuple represents a subject and its corresponding mark.

    Returns:
        A new list of tuples, sorted by the second value of each tuple.
    """
    return sorted(marks
\end{lstlisting}

\textbf{Baseline Generation:}

\begin{lstlisting}
, key=lambda x: x[1])
\end{lstlisting}

\textbf{Steering Token and Coefficient:} \texttt{`1'}, \texttt{Coeff -80}

\textbf{Steered Continuation:}

\begin{lstlisting}
)
\end{lstlisting}

\subsection{Computing the $n$‑th tetrahedral number.}\label{sec:app_tetra}
\textbf{Prompt Prefix (up to token 180):}

\begin{lstlisting}
<bos>You are an expert Python programmer, and here is your task: Write a function to find the nth tetrahedral number. Your code should pass these tests:

assert tetrahedral_number(5) == 35
assert tetrahedral_number(6) == 56
assert tetrahedral_number(7) == 84
Write your code below starting with "```python" and ending with "```".
```python
def tetrahedral_number(n):
    """
    Calculates the nth tetrahedral number.

    Args:
        n: The position of the tetrahedral number in the sequence.

    Returns:
        The nth tetrahedral number.
    """
    if n <= 0:
        return 0
    elif n == 1:
        return 1
    else:
        return
\end{lstlisting}

\textbf{Baseline Generation:}

\begin{lstlisting}
 (n * (n + 1) * (n + 2)) // 6
\end{lstlisting}

\textbf{Steering Token and Coefficient:} \texttt{`1'}, \texttt{Coeff -80}

\textbf{Steered Continuation:}

\begin{lstlisting}
 tetrahedral_number(n-1) + 1
\end{lstlisting}

\subsection{Forming the maximum number from digits.}\label{sec:app_max_num}
\textbf{Prompt Prefix (up to token 311):}

\begin{lstlisting}
<bos>You are an expert Python programmer, and here is your task: Write a python function to find the largest number that can be formed with the given list of digits. Your code should pass these tests:

assert find_Max_Num([1,2,3]) == 321
assert find_Max_Num([4,5,6,1]) == 6541
assert find_Max_Num([1,2,3,9]) == 9321
Write your code below starting with "```python" and ending with "```".
```python
def find_Max_Num(digits):
    """
    Finds the largest number that can be formed with the given list of digits.

    Args:
        digits: A list of digits.

    Returns:
        The largest number that can be formed with the given digits.
    """
    \end{lstlisting}

\textbf{Baseline Generation:}

\begin{lstlisting}
digits.sort()
    n = len(digits)
    max_num = ''
    for i in range(n):
        max_num += str(digits[n-i-1])
    return int(max_num)
\end{lstlisting}

\textbf{Steering Token and Coefficient:} \texttt{`sort'}, \texttt{Coeff -60}

\textbf{Steered Continuation:}

\begin{lstlisting}
max_num = 0
    for num in digits:
        if num > max_num:
            max_num = num
    return max_num
\end{lstlisting}

%% file: appendices/cantsaycases.tex
\section{Details for ``can't say'' cases}
\label{appendix:cant-say}
This appendix presents three representative ``can't say'' cases. For each case, we show the same information as \S\ref{sec:add_cases}.

\subsection{Overlap with Prompt}
\label{sec:op_prompt}
\textbf{Prompt Prefix (up to token 311):}

\begin{lstlisting}
<bos>You are an expert Python programmer, and here is your task: Write a function to find tuples which have all elements divisible by k from the given list of tuples. Your code should pass these tests:

assert find_tuples([(6, 24, 12), (7, 9, 6), (12, 18, 21)], 6) == [(6, 24, 12)]
assert find_tuples([(5, 25, 30), (4, 2, 3), (7, 8, 9)], 5) == [(5, 25, 30)]
assert find_tuples([(7, 9, 16), (8, 16, 4), (19, 17, 18)], 4) == [(8, 16, 4)]
Write your code below starting with "```python" and ending with "```".
```python
def find_tuples(tuples_list, k):
    """
    Finds tuples in a list of tuples that have all elements divisible by k.

    Args:
        tuples_list: A list of tuples.
        k: The divisor.

    Returns:
        A list of tuples that have all elements divisible by k.
    """
    result = []
    for tup in tuples_list:
        if all(element % k
\end{lstlisting}

\textbf{Baseline Generation:}

\begin{lstlisting}
 == 0 for element in tup):
            result.append(tup)
    return result   
\end{lstlisting}

\textbf{Steering Token and Coefficient:} \texttt{`for'}, \texttt{Coeff -80}

\textbf{Steered Continuation:}

\begin{lstlisting}
, k):
            result.append(tup)
    return result
\end{lstlisting}

\subsection{Out-of-Distribution Steering}
\label{sec:ood_steer}
\textbf{Prompt Prefix (up to token 223):}

\begin{lstlisting}
<bos>You are an expert Python programmer, and here is your task: Write a python function to find the largest number that can be formed with the given list of digits. Your code should pass these tests:

assert find_Max_Num([1,2,3]) == 321
assert find_Max_Num([4,5,6,1]) == 6541
assert find_Max_Num([1,2,3,9]) == 9321
Write your code below starting with "```python" and ending with "```".
```python
def find_Max_Num(digits):
    """
    Finds the largest number that can be formed with the given list of digits.

    Args:
        digits: A list of digits.

    Returns:
        The largest number that can be formed with the given digits.
    """
    digits.sort()
    n = len(digits)
    max_num = ''
    for i in range(n):
        max_
\end{lstlisting}

\textbf{Baseline Generation:}

\begin{lstlisting}
num += str(digits[n-i-1])
    return int(max_num)
\end{lstlisting}

\textbf{Steering Token and Coefficient:} \texttt{`-'}, \texttt{Coeff -80}

\textbf{Steered Continuation:}

\begin{lstlisting}
digits = len(digits)
        """
    """
    """
    """
    """
    """
    """
    """
    """
    """
    """
    """
    """
    """
    """
    """
    """
    """
    """
    """
    """
    """
    """
    """
    """
    """
    """
    """
    """
    """
    """
\end{lstlisting}

\subsection{Both Overlap and Out-of-Distribution}
\label{sec:op_ood}
\textbf{Prompt Prefix (up to token 173):}

\begin{lstlisting}
<bos>You are an expert Python programmer, and here is your task: Write a function to find the nth tetrahedral number. Your code should pass these tests:

assert tetrahedral_number(5) == 35
assert tetrahedral_number(6) == 56
assert tetrahedral_number(7) == 84
Write your code below starting with "```python" and ending with "```".
```python
def tetrahedral_number(n):
    """
    Calculates the nth tetrahedral number.

    Args:
        n: The position of the tetrahedral number in the sequence.

    Returns:
        The nth tetrahedral number.
    """
    if n <= 0:
        return 0
    elif n == 1:
        return 1
\end{lstlisting}

\textbf{Baseline Generation:}

\begin{lstlisting}

    else:
        return (n * (n + 1) * (n + 2)) // 6
\end{lstlisting}

\textbf{Steering Token and Coefficient:} \texttt{`('}, \texttt{Coeff -100}

\textbf{Steered Continuation:}

\begin{lstlisting}
 the
    
        The
        The
        The
        The
        The
        The
        The
        The
        The
        The
        The
        The
        The
        The
        The
        The
        The
        The
        The
        The
        The
        The
        The
        The
        The
        The
        The
        The
        The
        The
        The
        The
\end{lstlisting}

%% file: appendices/ethics.tex
\section{Ethics Statement}
While planning detection has potential beneficial applications for enhancing model controllability, it also raises ethical concerns. The ability to manipulate model outputs could be misused to bypass safety measures or to make models generate harmful content. We emphasize the importance of responsible use of these techniques and suggest the development of countermeasures to protect against potential misuse.

%% file: appendices/artifacts.tex
\section{Artifacts}

\subsection{MBPP Dataset}

The original Mostly Basic Programming Problems (MBPP) dataset \citep{austin2021programsynthesislargelanguage} features 974 python programming problems featuring a text description of a function, along with a set of unit tests that a model's generated code is supposed to pass. We filtered this dataset down to the 120 tasks that Gemma-2-2B solves correctly with deterministic sampling (temperature 0). The entire subset used for all analysis; no train/dev/test split required since we perform interpretability on fixed generations.

\subsubsection{Tasks}

\textbf{Compliance} This dataset uses the Creative Commons Attribution 4.0 International (CC BY 4.0) license.

\subsection{Gemma-2-2B}
We use Gemma-2-2B (2 billion parameters) – a decoder-only Transformer with 26 layers and RMSNorm pre- and post-normalization (see Team et al., 2024 for full architecture and training details).

\textbf{Compliance} Gemma-2 models are released under Google’s commercially-friendly Gemma License, which permits model usage for research and evaluation purposes only. 

\subsection{Sparse Autoencoders}
We used code and sparse autoencoder weights (SAE) from the GemmaScope release, trained on the base Gemma-2-2B model. We 26 SAEs, one for each layer, trained to reconstruct the outputs of the MLP layers, before the post-RMSNorm is applied. These SAEs use the TopK activation function with $K=32$, a latent dimension of $2048$ matching the MLP out dimension, and have an expansion factor of $8$ for a total of $16384$ features per layer.

\textbf{Compliance} The code and model checkpoints are distributed freely under the Apache 2.0 license.

%% file: appendices/compute.tex
\section{Experimental Details}
\textbf{Hardware and Compute} We used a single node of 4x NVIDIA A40 GPUs (48 GB VRAM). Total compute ~250 GPU-hours across all experiments: Computing Sparse Feature Circuits using attribution patching took $10$ hours to run across all prompts and token generations. Computing precursor influence by steering clusters of features took the bulk of the compute with about 240 GPU-hours total.

\textbf{Algorithm Hyperparameters}
We based our implementation for attribution patching from the source code from \citep{marks2025sparsefeaturecircuitsdiscovering}, code available \href{https://github.com/saprmarks/feature-circuits}{\underline{here}}. The original algorithm uses clean and counterfactual pairs of prompts, whereas we use only clean prompts then perform zero ablations on intermediate representations.

The algorithm takes a metric $m$ to backpropagate through, a hyperparemeter $\tau$ for the metric, and number of integrated gradient steps $n$. We use the probability of the correct token $p(y_\text{correct})$ as the metric. We set $\tau=0.60$, meaning we keep nodes that preserve the correct token's probability to be above 60\%, and set $n=10$.

%% file: appendices/aistatement.tex
\section{Statement on the Usage of Generative AI}
We used generative AI tools (e.g., GitHub Copilot and ChatGPT) to streamline routine coding tasks—such as writing data‐loading scripts. In each case, all AI‐suggested code was carefully reviewed, tested, and revised by the authors to ensure correctness and maintain consistent coding style. We used ChatGPT with search to generate high‐level literature summaries that informed our reference list and contextual background, which were cross-checked against original papers before inclusion.